%% file: main.tex
\documentclass{article} 
\usepackage{iclr2027_conference,times}

\iclrfinalcopy

\input{math_commands.tex}

\usepackage{amsthm}

\usepackage{amsmath}
\usepackage{amssymb}
\usepackage{algorithm}
\usepackage{algpseudocode}
\usepackage{booktabs}
\usepackage{float}
\usepackage{hyperref}
\usepackage{url}
\usepackage{booktabs}
\usepackage{tabularx}
\usepackage{wrapfig}
\usepackage{multirow}
\usepackage{rotating}
\usepackage{graphicx}
\usepackage{microtype}
\usepackage{xcolor}
\newcommand{\blue}[1]{{{\color{blue} #1}}}

\usepackage[capitalize,noabbrev]{cleveref}
\usepackage[normalem]{ulem}

\definecolor{myblue}{HTML}{598BE7}

\title{Decoupling Policy Extraction for Offline Reinforcement Learning}

\author{Xuyao Lin\thanks{Equal Contribution.} \\
Simplexity Robotics, Rensselaer Polytechnic Institute\\
\texttt{linxuyao@s-robots.com} \\
\AND 
Yixiang Shan$^{*}$ \\
Simplexity Robotics\\
\texttt{shanyixiang@s-robots.com} \\
\AND 
Jinru Duan \\
Simplexity Robotics, Northeastern University\\
\texttt{duanjinru@s-robots.com} \\
\AND
Tao Yang, Xinyu Zhao, Runyu Lei, Yiming Zhao, Jiaxin Fan, Zongbao Feng\thanks{Corresponding Author.}~~~\& Peng Jia \\
Simplexity Robotics \\
\texttt{\{yangtao, zhaoxinyu, leirunyu, -, fanjiaxin, fengzongbao, jiapeng\}} \\
\texttt{@s-robots.com}
}

\begin{document}

\maketitle

\fancyhead[L]{\small Preprint.}

\input{sections/1.abstraction}

\input{sections/2.introduction}

\input{sections/3.preliminary}
\input{sections/pre4.discussion}

\input{sections/4.methods}

\input{sections/5.experiments}

\input{sections/6.relatedworks}

\input{sections/7.conclusion}



\bibliography{iclr2027_conference}
\bibliographystyle{iclr2027_conference}

\appendix
\input{sections/8.appendix}

\end{document}

%% file: math_commands.tex
\usepackage{amsmath,amsfonts,bm}

\def\eqref#1{equation~\ref{#1}}

\def\1{\bm{1}}

\DeclareMathAlphabet{\mathsfit}{\encodingdefault}{\sfdefault}{m}{sl}
\SetMathAlphabet{\mathsfit}{bold}{\encodingdefault}{\sfdefault}{bx}{n}



%% file: sections/1.abstraction.tex

\begin{abstract}
Offline RL methods commonly jointly train the actor and critic, where the critic is used to guide the actor toward higher-value actions. This coupled learning process is well motivated in online RL, where an improved actor collects new data that can further update the actor and the critic.
However, training data remains fixed in offline RL, making actor-side policy improvement
unable to generate new data to validate or correct the critic. 
Moreover, retaining this coupled paradigm leads to two related challenges. Firstly, actor updates can drift toward high-valued but potentially out-of-distribution (OOD) actions and amplify critic overestimation. Secondly, conservative value estimation or behavior-cloning regularization creates a difficult trade-off between suppressing OOD actions and selecting high-value actions within the data-supported region.
Motivated by this observation, we revisit the conventional offline RL paradigm and propose decoupling policy improvement from actor training.
Specifically, we train the actor solely to model the behavior distribution and perform policy improvement at inference time by reranking multiple actor-generated proposals with a separately learned critic.
We refer to this paradigm as the decoupled policy extraction paradigm.
Under such paradigm, the actor provides behavior-supported action candidates, while the critic performs value-based selection within this candidate set.
Extensive experiments show that the decoupled policy extraction paradigm outperforms both behavior cloning and jointly learned offline RL methods, while remaining effective even with a naive Q-learning critic.
\end{abstract}

%% file: sections/2.introduction.tex
\section{Introduction}
\label{sec:introduction}


Reinforcement learning (RL) learns decision-making policies from reward
feedback, {which is} commonly implemented
with an actor--critic architecture. The critic estimates the long-term value of actions, while the actor is trained under its guidance to favor better actions.
This division of labor is especially natural in online RL, because the critic drives the actor toward high-value actions, while the improved actor in turn expands the critic's data support and corrects its estimation errors through new environment interactions, thereby forming a meaningful closed-loop training process.


\begin{wrapfigure}{r}{0.48\textwidth}
    \centering
    \includegraphics[width=\linewidth]{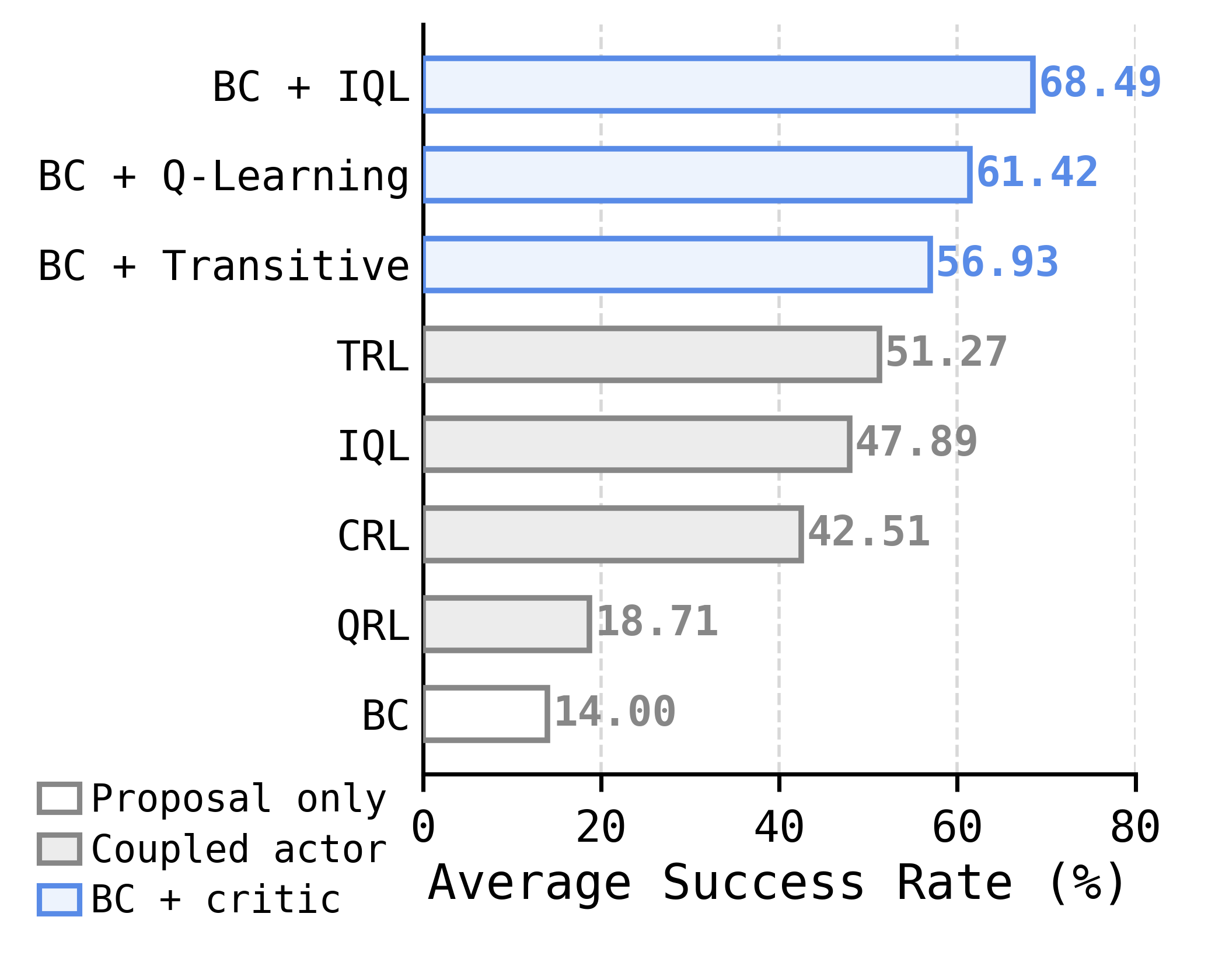}
    \caption{
        Aggregate performance on 30 offline GCRL tasks. Decoupled policy extraction variants outperforms original coupled baselines. Full task-level results and hyperparameters are provided in \cref{sec:appendix:crl-qrl}.
    }
    \label{fig:intro}
\end{wrapfigure}

Although a broad class of offline RL methods has retained such an actor--critic structure~\citep{fujimoto2019bcq,kumar2020cql,kostrikov2022iql}, we believe the rationale for retaining actor-side policy improvement is substantially weaker in offline RL. 
As the dataset is fixed, improving the actor neither changes the training distribution nor collects new experiences that could correct critic errors. 
Moreover, retaining this training paradigm suffers from two fundamental issues: the \textbf{OOD amplification loop} problem and the \textbf{support–value trade-off} problem.
Firstly, the actor continually shifts toward potentially out-of-distribution actions assigned higher values by the critic, thereby amplifying the effect of critic overestimation and further shifting the policy away from the data distribution, leading to the OOD amplification loop problem.
Secondly, existing methods typically introduce conservative value estimation or behavior cloning regularization to control this deviation. However, weak constraints may fail to suppress OOD actions, whereas strong constraints may prevent the policy from selecting high-value actions even within the data-supported region, resulting in a difficult trade-off between value maximization and distributional support, namely the support–value trade-off problem.

Building on the above observation, we revisit the paradigm design for offline reinforcement learning by decoupling policy improvement from actor learning. Specifically, we propose to learn a behavior cloning actor and learn a critic independently. At inference time, the actor proposes multiple supported candidate actions, which are reranked and selected by the critic. 
In contrast to prior coupled policy improvement methods, we refer to this paradigm as \textbf{decoupled policy extraction paradigm}. Under this paradigm, policy improvement is no longer implicitly realized through critic-driven actor optimization, but is instead explicitly instantiated as test-time action selection. 
This decoupling naturally addresses the two issues discussed above. First, because the critic no longer updates the actor, critic overestimation is not recursively amplified through actor optimization, thereby avoiding the OOD amplification loop. Second, distributional support and value maximization are no longer balanced within a single actor objective, avoiding the support–value trade-off inherent in coupled policy improvement.

The aggregate results in \cref{fig:intro} demonstrate the advantage of decoupling policy improvement from actor learning. Across 30 offline goal-conditioned tasks, replacing the original critic-guided actor with a decoupled policy extraction variant consistently improves performance. For example, with the IQL critic, the decoupled variant increases the average success rate from (47.89\%) to (68.49\%). Similar improvements with other critic backbones indicate that the benefit is not specific to a particular value-learning objective. These results support the decoupled policy extraction paradigm, in which the actor is responsible only for modeling the data-supported action distribution, while policy improvement is performed explicitly through critic-guided selection at inference time.
It is worth noting that although related inference-time policy optimization approaches have appeared in several existing methods~\citep{hansenestruch2023idql, nakamoto2024vgps}, we aim to provide a systematic discussion of RL method mechanisms. The relationship between our paradigm and existing methods will be discussed in \cref{sec:related_work}.

Our contributions are as follows:
\begin{itemize}
    \item 
    We revisit the prevailing coupled policy improvement paradigm in offline RL and identify its limitations. Based on this analysis, we propose a decoupled policy extraction paradigm which suits offline setting better.

    \item We instantiate this paradigm with multiple critic backbones and show that decoupled policy extraction consistently improves performance over the corresponding coupled methods across a broad range of offline goal-conditioned tasks. 
    \item 
    The proposed paradigm has important implications for large-scale policies. By delegating policy improvement to a lightweight critic while avoiding repeated updates to the actor, it provides a more computationally efficient approach to offline RL with large policy models.
\end{itemize}




%% file: sections/3.preliminary.tex
\section{Preliminaries}
\label{sec:preliminaries}

\subsection{Offline Goal-Conditioned Reinforcement Learning}
\label{subsec:offline_gcrl}

We consider a goal-conditioned Markov decision process
\[
    \mathcal M=(\mathcal S,\mathcal A,\mathcal G,P,r,\rho_0,p_G,\gamma,H),
\]
where \(s\in\mathcal S\), \(a\in\mathcal A\), and \(g\in\mathcal G\) denote the
state, action, and goal. A goal-conditioned policy \(\pi(a\mid s,g)\) maximizes
\[
    J(\pi)
    =
    \mathbb E_{g\sim p_G,\;\tau\sim p^\pi(\cdot\mid g)}
    \left[
        \sum_{t=0}^{T-1}\gamma^t r(s_t,a_t,g)
    \right],
    \qquad T\le H .
\]

We write \(x=(s,g)\) for a state--goal context. The corresponding action-value
function is
\[
    Q^\pi(x,a)
    =
    \mathbb E_\pi
    \left[
        \sum_{k=0}^{T-t-1}\gamma^k r(s_{t+k},a_{t+k},g)
        \mid x_t=x,a_t=a
    \right].
\]

In offline RL, the learner has access only to a fixed dataset
\[
    \mathcal D=\{(s_i,a_i,r_i,s_i',d_i,g_i)\}_{i=1}^{M},
\]
and cannot collect additional interaction data. The dataset induces a context
marginal \(\rho_{\mathcal D}(x)\) and a behavioral conditional action
distribution \(\mu_{\mathcal D}(a\mid x)\). Since value estimates are most
reliable on state--action regions covered by the dataset, optimizing a policy
toward weakly supported actions may exploit extrapolation errors that cannot be
corrected through new interaction. We therefore use the behavior-supported
region as the reference for discussing action coverage and out-of-distribution
drift.

\subsection{Value Learning and Policy Extraction}
\label{subsec:value_and_extraction}

We separate critic learning from policy extraction. Let \(c\) denote a
critic-learning rule, such as Q-Learning, IQL, or TRL. We write
\[
    \widehat Q^c = \mathsf{ValueLearn}_c(\mathcal D),
    \qquad
    \Pi = \mathsf{Extract}(\mathcal D,\widehat Q^c),
\]
where \(\widehat Q^c(x,a)\) is the scalar score used for action selection,
regardless of auxiliary value functions or target networks used during critic
training.

For selecting among a finite candidate set, only the relative ordering of
actions matters:
\[
    \arg\max_{a\in\mathcal C} h(\widehat Q^c(x,a))
    =
    \arg\max_{a\in\mathcal C} \widehat Q^c(x,a)
\]
for any strictly increasing function \(h\). Thus, different critics can be
compared through a shared ranking-based extraction interface.

Many offline RL methods extract policies by training an actor with
critic-derived supervision. For example, regularized actor objectives optimize
high-value actions while penalizing deviation from the behavior distribution,
and advantage-weighted regression fits dataset actions with weights derived from
critic estimates. We refer to any actor whose training objective depends on
critic values, advantages, or other critic-derived signals as an
\emph{RL-trained actor}. In contrast, a \emph{behavior-cloned proposer}
\(q_{\mathrm{BC}}(a\mid x)\) is trained only from dataset actions and receives
no critic-derived supervision.

\subsection{Proposal-and-Selection Policies}
\label{subsec:proposal_selection_prelim}

We use \(q(a\mid x)\) to denote a stochastic action proposer. Given a context
\(x\), the proposer samples \(N\) candidate actions
\[
    C_N(x;q)=\{a_1,\ldots,a_N\},
    \qquad
    a_i\sim q(\cdot\mid x).
\]
A critic-based selector chooses
\[
    \Pi[q,\widehat Q^c,N](x)
    =
    \arg\max_{a_i\in C_N(x;q)}
    \widehat Q^c(x,a_i).
\]
As a control, we also consider random selection from the same candidate set,
\[
    \Pi_{\mathrm{rand}}[q,N](x)
    \sim
    \mathrm{Unif}(C_N(x;q)),
\]
which measures the effect of sampling multiple actions without using critic
ranking. When \(N=1\), no comparison is performed and the deployed action is a
single proposal. When \(N>1\), critic-based selection changes the deployed
action distribution at inference time without retraining the proposer.

This notation makes three factors explicit: the proposal distribution \(q\), the
critic ranking function \(\widehat Q^c\), and the candidate budget \(N\). In our
experiments, we instantiate \(q\) as a frozen behavior-cloned proposer and vary
the critic backbone and candidate budget to diagnose offline policy extraction.
In distributional diagnostics, we use the behavior-cloned action distribution as
an empirical proxy for the behavior-supported region.

%% file: sections/pre4.discussion.tex
\section{Diagnosing Coupled Policy Improvement in Offline RL}
\label{sec:exp:failure analysis}

As discussed in \cref{sec:introduction}, jointly trained actor--critic methods are subject to two key limitations: the amplification of critic errors through the actor--critic feedback loop, and the support--value trade-off. In this section, we empirically investigate these two issues separately. We further test whether decoupling proposal learning from critic-based selection
changes performance under compute-matched candidate search with a fixed critic.
Detailed settings are provided in Appendices~\ref{sec:appendix:actor-drift-hparams} and \ref{sec:appendix:cql-tradeoff}.




\subsection{Out-of-distribution Amplification Loop}
\begin{figure}[!htbp]
    \centering
    \includegraphics[
        width=\linewidth,
        height=0.9\textheight,
        keepaspectratio
    ]{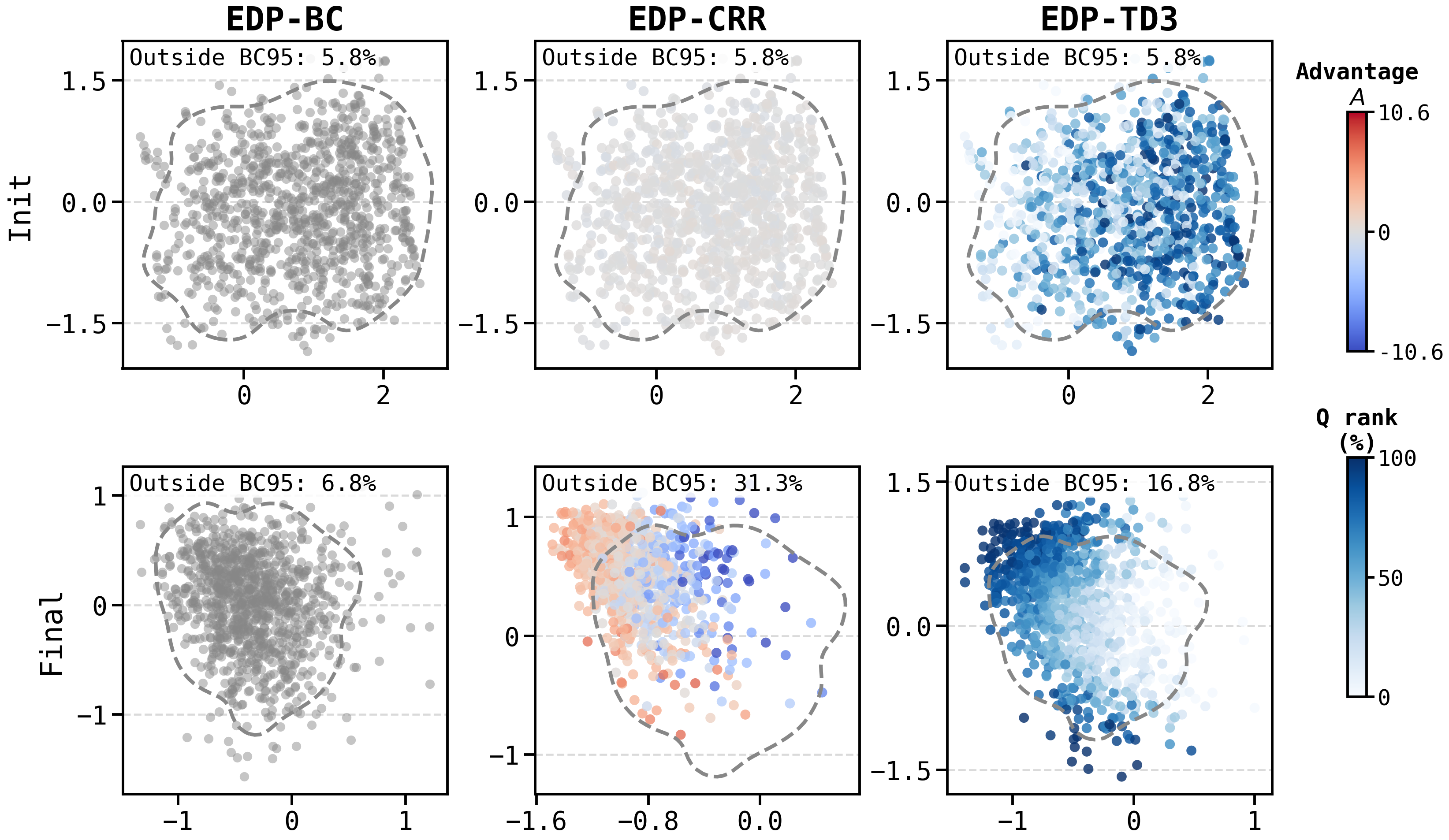}
    \caption{Dashed contours mark the same-checkpoint 95\% conformal BC reference regions,
  with outside rates reported in each panel.
  Colors encode raw diagnostic advantages for CRR and within-checkpoint
  \(\min(Q_1,Q_2)\) percentile ranks for TD3.
    }
    \label{fig:exp:distribution_shift}
\end{figure}
\label{sec:distribution_shift}

{
Out-of-distribution (OOD) actions have been recognized as a central challenge in offline reinforcement learning. 
A key source of this difficulty lies in the interaction between critic estimation error and coupled policy improvement. 
When the actor assigns probability mass to actions outside the support of the offline dataset, the critic may consequently overestimate some of them. 
Directly optimizing the actor against such estimates can then increase the probability of these overestimated OOD actions, further shifting the actor toward regions where the critic is less reliable. This creates a self-reinforcing feedback loop between critic error and actor distribution shift. 
In online RL, such errors can be corrected because the updated policy generates new interactions that provide training data for these newly visited regions. In offline RL, however, the dataset remains fixed, leaving this feedback loop without an equivalent correction mechanism.
}

We next empirically examine this feedback-loop error amplification in coupled policy improvement methods without explicit pessimistic regularization.
\Cref{fig:exp:distribution_shift} compares EDP-BC, EDP-CRR and
 EDP-TD3, representing BC policy, advantage-weighted regression, and behavior-regularized
  \(Q\) maximization, respectively. 
  Although all methods use the same initialization, after training, the fraction
  of actions outside the same-checkpoint BC95 region differs;
  the fraction is \(6.8\%\) for EDP-BC,
  compared with \(31.3\%\) for EDP-CRR and \(16.8\%\) for EDP-TD3.
  These results highlight a fundamental vulnerability of coupled policy improvement: directly optimizing the actor against an imperfect offline critic can amplify distribution shift through the actor--critic feedback loop.
  Moreover,
  actions beyond BC95 receive higher diagnostic advantages under CRR and higher
  within-checkpoint \(Q\) ranks under TD3. These results provide evidence of
  critic-aligned actor drift: coupled updates can move the actor beyond the BC
  reference support, whereas freezing the proposal policy removes this feedback
  path. 
\subsection{The Support–Value Trade-off}

\label{sec:exp:conservatism-tradeoff}
{As discussed earlier, under the coupled policy improvement paradigm, out-of-distribution actions can induce a looped error amplification between critic estimation errors and actor updates, thereby leading to actor distribution shift. Existing offline RL methods commonly address this issue through conservative or behavior-cloning regularization. However, we believe that these mechanisms do not eliminate the underlying conflict; instead, they trade off policy improvement against adherence to the support of the offline dataset.} 

\begin{wrapfigure}{!h}{0.50\textwidth}
    \centering
    \includegraphics[width=1.0\linewidth]{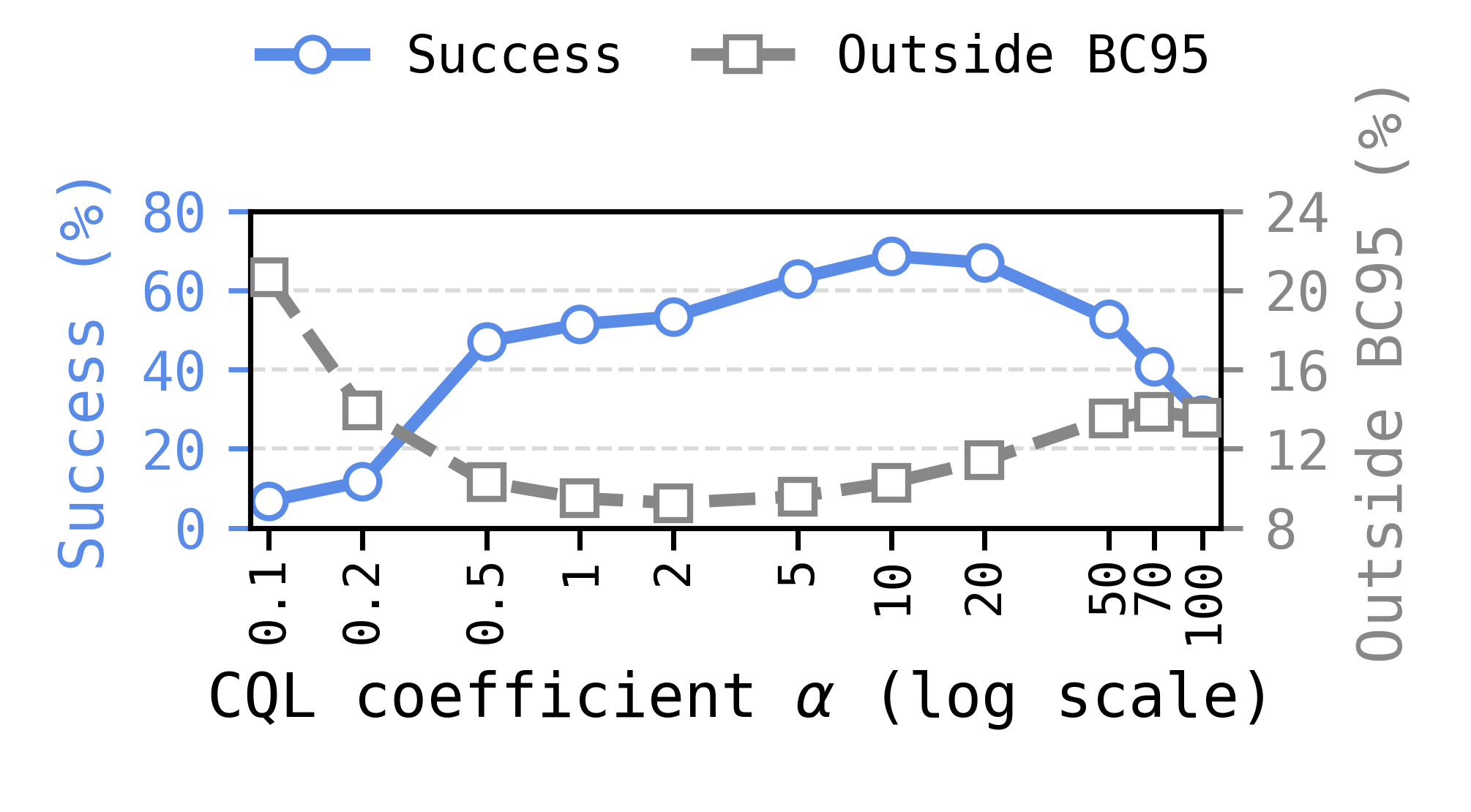}
    \caption{Observed success and BC support across selected CQL conservatism settings
      on Cube Double.
      The coefficient that maximizes success differs from that minimizing
      Outside BC95, while stronger pessimism eventually reduces success without
      improving the support diagnostic.}
    \label{fig:exp:claim2}
\end{wrapfigure}

Taking CQL as an example, we examine goal-conditioned CQL~\citep{kumar2020cql}, where
  $\alpha_{\mathrm{CQL}}$ scales the conservative critic penalty while the
  SAC actor objective remains unchanged.
  \Cref{fig:exp:claim2} reports task success together with the fraction
  of sampled actor actions outside the BC95 reference region.
As can be observed, as $\alpha$ increases, the ratio of actions outside BC95 initially decreases, with the performance of CQL improves. 
This is because that a moderately stronger conservative penalty effectively suppresses the Q-values of out-of-distribution actions, thereby reducing their likelihood of being selected by the policy. 
However, the two metrics reach their optima at different values of $\alpha$: the outside-BC95 ratio reaches its minimum at $\alpha=2$, whereas the success rate peaks at $\alpha=10$. 
This discrepancy indicates that the value of $\alpha$ that best constrains the policy to the behavior distribution does not necessarily maximize task performance. 
In other words, pessimistic methods such as CQL exhibit an inherent support--value trade-off: stronger conservatism can improve distributional support, but the degree of conservatism that yields the strongest support constraint is not necessarily the one that produces the best value-based decision making.

Interestingly, when $\alpha$ is increased further, the outside-BC95 ratio no longer decreases monotonically, while performance begins to deteriorate. We attribute this behavior to excessive conservatism. When the conservative penalty becomes too strong, it may suppress Q-values broadly rather than selectively penalizing unsupported actions, compressing the dynamic range of the learned value function. As a result, the critic becomes less discriminative among candidate actions, including between well-supported and poorly supported ones, which can ultimately degrade policy performance. 


%% file: sections/4.methods.tex
\section{Decoupled Policy Extraction}
\label{sec:method}

Building on the
notation in \cref{subsec:value_and_extraction,subsec:proposal_selection_prelim},
we instantiate policy extraction with a frozen BC actor and a
separately learned critic. The BC actor models dataset distribution and propose data-supported candidate actions, 
while the critic is used only to rank candidates at deployment time.

Unlike previous actor-side policy improvements, where critic-derived information changes the actor parameters during training, our decoupled policy extraction paradigm applies critic information only at inference time, thereby avoiding the aforementioned OOD amplification loop problem and the support-value trade-off problem.


\subsection{Frozen BC actor}
\label{subsec:bc_proposal}

We train the actor by behavior cloning. Given the offline dataset
\(\mathcal D\), the actor \(q_\psi(a\mid x)\) is optimized by maximum the log-likelihood:
\[
    \mathcal L_{\mathrm{BC}}(\psi)
    =
    -
    \mathbb E_{(x,a)\sim\mathcal D}
    \left[
        \log q_\psi(a\mid x)
    \right].
    \label{eq:bc_proposal_objective}
\]
For deterministic parameterizations, this reduces to supervised action
regression.

After training, the actor is frozen. It receives no rewards, critic values,
advantage weights, or gradients from \(\widehat Q^c\). Thus, any improvement over
behavior cloning comes from critic-based selection over fixed actor samples,
rather than from critic-guided updates to the actor. This also prevents
critic-derived preferences from changing the proposal distribution during
training, as discussed in 
\cref{sec:exp:failure analysis}.

\subsection{choice of critic}

One of the advantage of decoupled policy extraction paradigm is that it substantially relaxes the requirements on critic design. In conventional offline RL, the critic is often explicitly regularized or conservatively trained to avoid assigning spuriously high values to OOD actions, since such overestimation can directly drive the actor toward unsupported regions. In our framework, however, policy improvement is performed by reranking actions proposed by a BC actor. Since the BC actor approximates the dataset distribution, the candidate set is largely restricted to data-supported regions. 

As a result, the critic is no longer required to explicitly suppress arbitrary OOD actions over the entire action space, but only to provide sufficiently accurate relative value estimates among behavior-supported candidates. This substantially reduces the need for complex conservative critic designs and alleviates the support--value trade-off discussed in \cref{sec:exp:conservatism-tradeoff}. Consequently, our framework can accommodate a broad range of critic objectives, including naive value-learning methods that do not explicitly account for OOD actions. We further show in \cref{sec:exp:main-results} that even a vanilla Q-learning critic can achieve strong performance under our framework.

\subsection{policy extraction via reranking}
\label{subsec:backbone_and_budget}



Our policy extraction is formulated as a reranking process: the BC actor proposes $N$ data-supported action candidates, which are evaluated by the critic, and the highest-valued action is selected for execution.
The candidate budget \(N\) controls the amount of inference-time search within
the proposal distribution. Larger \(N\) gives the selector more candidates to
compare, but can also increase exposure to ranking errors or weakly supported actions.

Therefore, we effectively replaces most of the complex OOD-control machinery in coupled policy improvement methods with the tuning of a single inference-time parameter, the candidate budget $N$. In coupled methods, mitigating OOD overestimation requires either redesigning the method or repeatedly tuning the coefficient that balances behavioral support against value maximization. Both procedures involve costly training-time iterations.
In contrast, the proposed decoupled policy extraction paradigm shifts this expensive model-level iteration to a lightweight inference-time selection problem. The candidate budget $N$ can be adjusted without retraining either the actor or the critic, substantially reducing the cost of policy iteration.




%% file: sections/5.experiments.tex
\section{Experiments}
\label{subsec:evaluation_protocol}

In this section, we conduct experiments to evaluate the performance of our proposed method and analyze the impact of its key design.
The computational resources and hyper-parameters details are available in \cref{sec:appendix:Implementation Details}.


\subsection{Experimental Setup}

\paragraph{Benchmarks.}
We evaluate six goal-conditioned continuous-control environments from
OGBench \citep{park2025ogbench}: AntMaze-Large, AntSoccer-Arena, Cube-Double, HumanoidMaze-Medium, Puzzle-4x4, and
Scene-Play.  These domains span locomotion, object manipulation, contact-rich
control, and state dimensions for which a single unimodal proposal can be
restrictive.  Each environment contributes its five fixed benchmark tasks.

\paragraph{Backbones.}
We select Q-Learning, IQL, and TRL as representative backbones to evaluate our proposition. Q-Learning represents the most fundamental Bellman-based value learning method. IQL learns the state value function through expectile regression, enabling policy improvement without explicitly maximizing over out-of-distribution actions, and has become a widely adopted offline reinforcement learning method. TRL accelerates long-horizon value learning by exploiting value transitivity and the triangle inequality. 
For all backbones, we retain their original critic learning mechanisms while uniformly using BC to generate candidate actions and the learned critic to rerank them.


\paragraph{Training and evaluation.}
  For each frozen-proposal method, we first optimize FM-BC for \(10^6\) gradient
  steps, freeze its parameters, and then optimize the critic for \(10^6\) steps.
  FM-IQL and TRL-RPG jointly optimize their actor and critic for \(10^6\) steps.
  For the standard benchmark, we average the evaluations at 800K, 900K, and 1M
  updates within each training seed and then report the mean and standard
  deviation across seeds. Candidate-budget sweeps are evaluated separately at
  the 1M checkpoint. Full seed coverage and implementation details are provided
  in \cref{sec:appendix:hyperparameters}.

\subsection{Main Results}
\label{sec:exp:main-results}

\begin{table}[!th]
\centering
\caption{OGBench success rates (\%; mean $\pm$ s.d.\ over five seeds when
reported).}
\label{tab:overall-results}

\vspace{0.8em}

\resizebox{0.9\linewidth}{!}{
\begin{tabular}{@{}llllllll@{}}
\toprule
Dataset & Task & FM-BC & FM-IQL & TRL
& \multicolumn{3}{c}{BC + critic reranking} \\
\cmidrule(lr){6-8}
& & & & & IQL & Transitive & Q-Learning \\
\midrule

\multirow{6}{*}{AntMaze-Large}
& task1
& 22 {\scriptsize $\pm$ 10}
& 32 {\scriptsize $\pm$ 4}
& 52 {\scriptsize $\pm$ 8}
& 59 {\scriptsize $\pm$ 17}
& \textcolor{myblue}{73} {\scriptsize $\pm$ 7}
& 28 {\scriptsize $\pm$ 22} \\

& task2
& 24 {\scriptsize $\pm$ 3}
& 29 {\scriptsize $\pm$ 7}
& \textcolor{myblue}{52} {\scriptsize $\pm$ 12}
& 11 {\scriptsize $\pm$ 3}
& 27 {\scriptsize $\pm$ 11}
& 23 {\scriptsize $\pm$ 7} \\

& task3
& 49 {\scriptsize $\pm$ 10}
& 73 {\scriptsize $\pm$ 9}
& 81 {\scriptsize $\pm$ 9}
& 81 {\scriptsize $\pm$ 7}
& \textcolor{myblue}{85} {\scriptsize $\pm$ 8}
& 67 {\scriptsize $\pm$ 9} \\

& task4
& 12 {\scriptsize $\pm$ 9}
& 10 {\scriptsize $\pm$ 4}
& 19 {\scriptsize $\pm$ 2}
& 5 {\scriptsize $\pm$ 5}
& 19 {\scriptsize $\pm$ 19}
& \textcolor{myblue}{23} {\scriptsize $\pm$ 9} \\

& task5
& 17 {\scriptsize $\pm$ 3}
& 12 {\scriptsize $\pm$ 6}
& 27 {\scriptsize $\pm$ 7}
& 15 {\scriptsize $\pm$ 8}
& 39 {\scriptsize $\pm$ 11}
& \textcolor{myblue}{55} {\scriptsize $\pm$ 13} \\

& \textbf{overall}
& 25 {\scriptsize $\pm$ 3}
& 31 {\scriptsize $\pm$ 3}
& 46 {\scriptsize $\pm$ 3}
& 34 {\scriptsize $\pm$ 2}
& \textcolor{myblue}{49} {\scriptsize $\pm$ 4}
& 41 {\scriptsize $\pm$ 6} \\

\midrule

\multirow{6}{*}{HumanoidMaze-Medium}
& task1
& 4 {\scriptsize $\pm$ 3}
& 33 {\scriptsize $\pm$ 16}
& 75 {\scriptsize $\pm$ 6}
& \textcolor{myblue}{81} {\scriptsize $\pm$ 13}
& 77 {\scriptsize $\pm$ 14}
& 76 {\scriptsize $\pm$ 16} \\

& task2
& 8 {\scriptsize $\pm$ 3}
& 55 {\scriptsize $\pm$ 9}
& 92 {\scriptsize $\pm$ 4}
& \textcolor{myblue}{93} {\scriptsize $\pm$ 7}
& 89 {\scriptsize $\pm$ 5}
& 88 {\scriptsize $\pm$ 8} \\

& task3
& 13 {\scriptsize $\pm$ 6}
& 1 {\scriptsize $\pm$ 1}
& 3 {\scriptsize $\pm$ 2}
& 19 {\scriptsize $\pm$ 12}
& \textcolor{myblue}{40} {\scriptsize $\pm$ 11}
& 0 {\scriptsize $\pm$ 0} \\

& task4
& 2 {\scriptsize $\pm$ 1}
& 0 {\scriptsize $\pm$ 0}
& \textcolor{myblue}{17} {\scriptsize $\pm$ 8}
& 0 {\scriptsize $\pm$ 0}
& 5 {\scriptsize $\pm$ 5}
& 0 {\scriptsize $\pm$ 0} \\

& task5
& 8 {\scriptsize $\pm$ 3}
& 4 {\scriptsize $\pm$ 3}
& 81 {\scriptsize $\pm$ 10}
& 57 {\scriptsize $\pm$ 16}
& \textcolor{myblue}{92} {\scriptsize $\pm$ 8}
& 0 {\scriptsize $\pm$ 0} \\

& \textbf{overall}
& 7 {\scriptsize $\pm$ 2}
& 19 {\scriptsize $\pm$ 3}
& 54 {\scriptsize $\pm$ 3}
& 50 {\scriptsize $\pm$ 5}
& \textcolor{myblue}{61} {\scriptsize $\pm$ 4}
& 33 {\scriptsize $\pm$ 4} \\

\midrule

\multirow{6}{*}{AntSoccer-Arena}
& task1
& 23 {\scriptsize $\pm$ 6}
& 88 {\scriptsize $\pm$ 4}
& 90 {\scriptsize $\pm$ 3}
& 95 {\scriptsize $\pm$ 8}
& \textcolor{myblue}{99} {\scriptsize $\pm$ 3}
& 95 {\scriptsize $\pm$ 5} \\

& task2
& 22 {\scriptsize $\pm$ 6}
& 85 {\scriptsize $\pm$ 8}
& 92 {\scriptsize $\pm$ 4}
& \textcolor{myblue}{99} {\scriptsize $\pm$ 3}
& 83 {\scriptsize $\pm$ 5}
& 96 {\scriptsize $\pm$ 5} \\

& task3
& 14 {\scriptsize $\pm$ 6}
& 90 {\scriptsize $\pm$ 5}
& 84 {\scriptsize $\pm$ 3}
& \textcolor{myblue}{96} {\scriptsize $\pm$ 5}
& 84 {\scriptsize $\pm$ 9}
& 83 {\scriptsize $\pm$ 12} \\

& task4
& 9 {\scriptsize $\pm$ 2}
& 45 {\scriptsize $\pm$ 4}
& 54 {\scriptsize $\pm$ 10}
& 81 {\scriptsize $\pm$ 8}
& 51 {\scriptsize $\pm$ 21}
& \textcolor{myblue}{83} {\scriptsize $\pm$ 3} \\

& task5
& 7 {\scriptsize $\pm$ 5}
& 83 {\scriptsize $\pm$ 3}
& 52 {\scriptsize $\pm$ 3}
& \textcolor{myblue}{95} {\scriptsize $\pm$ 5}
& 55 {\scriptsize $\pm$ 17}
& 89 {\scriptsize $\pm$ 9} \\

& \textbf{overall}
& 15 {\scriptsize $\pm$ 3}
& 78 {\scriptsize $\pm$ 2}
& 74 {\scriptsize $\pm$ 3}
& \textcolor{myblue}{93} {\scriptsize $\pm$ 2}
& 74 {\scriptsize $\pm$ 4}
& 89 {\scriptsize $\pm$ 3} \\

\midrule

\multirow{6}{*}{Cube-Double}
& task1
& 20 {\scriptsize $\pm$ 7}
& 90 {\scriptsize $\pm$ 4}
& 58 {\scriptsize $\pm$ 11}
& \textcolor{myblue}{96} {\scriptsize $\pm$ 5}
& 61 {\scriptsize $\pm$ 11}
& 83 {\scriptsize $\pm$ 9} \\

& task2
& 5 {\scriptsize $\pm$ 3}
& 90 {\scriptsize $\pm$ 6}
& 20 {\scriptsize $\pm$ 11}
& \textcolor{myblue}{100} {\scriptsize $\pm$ 0}
& 32 {\scriptsize $\pm$ 14}
& 77 {\scriptsize $\pm$ 8} \\

& task3
& 1 {\scriptsize $\pm$ 2}
& 85 {\scriptsize $\pm$ 15}
& 12 {\scriptsize $\pm$ 9}
& \textcolor{myblue}{100} {\scriptsize $\pm$ 0}
& 13 {\scriptsize $\pm$ 4}
& 63 {\scriptsize $\pm$ 7} \\

& task4
& 1 {\scriptsize $\pm$ 1}
& 16 {\scriptsize $\pm$ 4}
& 2 {\scriptsize $\pm$ 2}
& \textcolor{myblue}{33} {\scriptsize $\pm$ 13}
& 9 {\scriptsize $\pm$ 5}
& 12 {\scriptsize $\pm$ 8} \\

& task5
& 2 {\scriptsize $\pm$ 2}
& 56 {\scriptsize $\pm$ 16}
& 12 {\scriptsize $\pm$ 10}
& 65 {\scriptsize $\pm$ 11}
& 15 {\scriptsize $\pm$ 11}
& \textcolor{myblue}{71} {\scriptsize $\pm$ 9} \\

& \textbf{overall}
& 6 {\scriptsize $\pm$ 2}
& 67 {\scriptsize $\pm$ 4}
& 21 {\scriptsize $\pm$ 8}
& \textcolor{myblue}{79} {\scriptsize $\pm$ 4}
& 26 {\scriptsize $\pm$ 7}
& 61 {\scriptsize $\pm$ 5} \\

\midrule

\multirow{6}{*}{Scene-Play}
& task1
& 66 {\scriptsize $\pm$ 10}
& \textcolor{myblue}{100} {\scriptsize $\pm$ 0}
& 94 {\scriptsize $\pm$ 3}
& \textcolor{myblue}{100} {\scriptsize $\pm$ 0}
& \textcolor{myblue}{100} {\scriptsize $\pm$ 0}
& 85 {\scriptsize $\pm$ 13} \\

& task2
& 29 {\scriptsize $\pm$ 17}
& 93 {\scriptsize $\pm$ 5}
& 93 {\scriptsize $\pm$ 2}
& \textcolor{myblue}{100} {\scriptsize $\pm$ 0}
& 93 {\scriptsize $\pm$ 6}
& 76 {\scriptsize $\pm$ 12} \\

& task3
& 26 {\scriptsize $\pm$ 8}
& 99 {\scriptsize $\pm$ 1}
& 94 {\scriptsize $\pm$ 5}
& \textcolor{myblue}{99} {\scriptsize $\pm$ 3}
& 95 {\scriptsize $\pm$ 8}
& 92 {\scriptsize $\pm$ 8} \\

& task4
& 28 {\scriptsize $\pm$ 4}
& 27 {\scriptsize $\pm$ 7}
& 79 {\scriptsize $\pm$ 16}
& 37 {\scriptsize $\pm$ 31}
& \textcolor{myblue}{80} {\scriptsize $\pm$ 9}
& 24 {\scriptsize $\pm$ 18} \\

& task5
& 4 {\scriptsize $\pm$ 2}
& 1 {\scriptsize $\pm$ 1}
& \textcolor{myblue}{28} {\scriptsize $\pm$ 10}
& 19 {\scriptsize $\pm$ 9}
& 21 {\scriptsize $\pm$ 17}
& 13 {\scriptsize $\pm$ 14} \\

& \textbf{overall}
& 31 {\scriptsize $\pm$ 6}
& 64 {\scriptsize $\pm$ 2}
& \textcolor{myblue}{78} {\scriptsize $\pm$ 3}
& 71 {\scriptsize $\pm$ 8}
& \textcolor{myblue}{78} {\scriptsize $\pm$ 4}
& 58 {\scriptsize $\pm$ 11} \\

\midrule

\multirow{6}{*}{Puzzle-4x4}
& task1
& 0 {\scriptsize $\pm$ 0}
& 37 {\scriptsize $\pm$ 6}
& 44 {\scriptsize $\pm$ 10}
& 93 {\scriptsize $\pm$ 6}
& 65 {\scriptsize $\pm$ 11}
& \textcolor{myblue}{100} {\scriptsize $\pm$ 0} \\

& task2
& 0 {\scriptsize $\pm$ 0}
& 27 {\scriptsize $\pm$ 9}
& 24 {\scriptsize $\pm$ 8}
& 76 {\scriptsize $\pm$ 19}
& 43 {\scriptsize $\pm$ 9}
& \textcolor{myblue}{84} {\scriptsize $\pm$ 7} \\

& task3
& 1 {\scriptsize $\pm$ 1}
& 33 {\scriptsize $\pm$ 8}
& 39 {\scriptsize $\pm$ 7}
& \textcolor{myblue}{97} {\scriptsize $\pm$ 3}
& 73 {\scriptsize $\pm$ 15}
& 88 {\scriptsize $\pm$ 9} \\

& task4
& 0 {\scriptsize $\pm$ 1}
& 17 {\scriptsize $\pm$ 2}
& 35 {\scriptsize $\pm$ 6}
& 77 {\scriptsize $\pm$ 16}
& 45 {\scriptsize $\pm$ 15}
& \textcolor{myblue}{83} {\scriptsize $\pm$ 14} \\

& task5
& 2 {\scriptsize $\pm$ 2}
& 24 {\scriptsize $\pm$ 6}
& 33 {\scriptsize $\pm$ 5}
& 75 {\scriptsize $\pm$ 11}
& 44 {\scriptsize $\pm$ 14}
& \textcolor{myblue}{79} {\scriptsize $\pm$ 9} \\

& \textbf{overall}
& 1 {\scriptsize $\pm$ 1}
& 28 {\scriptsize $\pm$ 4}
& 35 {\scriptsize $\pm$ 4}
& 84 {\scriptsize $\pm$ 7}
& 54 {\scriptsize $\pm$ 6}
& \textcolor{myblue}{87} {\scriptsize $\pm$ 6} \\

\bottomrule
\end{tabular}
}

\end{table}

Aggregate results are shown in \cref{tab:overall-results}.
As shown in the table, using IQL as an example, the decoupled approach yields relative improvements of 38.4\%, 13.9\%, and 202.2\% in macro-average success rate across navigation, manipulation, and logical-reasoning tasks, respectively. Taken together, these results empirically validate the effectiveness of decoupling policy improvement from actor training.

It is also worth noting that even a naive Q-learning method can serve as an effective critic for reranking. Compared with the widely adopted IQL, the decoupled paradigm based on Q-Learning achieves clear performance gains in most environments. 
The improvement is particularly pronounced in Puzzle, where reranking with Q-Learning achieves performance improvement of 212.8\% compared with IQL. We attribute this gap to the conservative expectile value-learning of IQL. 
By replacing hard maximization with upper-expectile regression over in-dataset actions, IQL avoids querying potentially out-of-distribution actions but weakens the propagation of rare in-distribution high-return continuations. This effect can accumulate in long-horizon sparse-reward tasks and compress the value gap between high-quality and ordinary candidates, thereby reducing reranking accuracy. For further analysis, please refer to \cref{sec:exp:failure analysis}.

\subsection{Ablation Analysis of Performance Gains}
\label{sec:exp:matched-proposer}
\begin{table}[]
\centering
\caption{Overall performance of FM-IQL variants on Puzzle-4x4 across different candidate budgets $N$.
}
\label{tab:exp:matched-proposer}
\resizebox{0.9\linewidth}{!}{
\begin{tabular}{@{}lllllll@{}}
\toprule
\textbf{FM-IQL Variants}     & {N = 1} & {N = 2} & {N = 4} & {N = 8} & {N = 16} & {N = 32} \\ \midrule
Coupled (Original)  & \underline{32.0±3.9} & N/A          & N/A           & N/A        & N/A         & N/A         \\
Coupled + Reranking          & \underline{32.0±4.7} & \underline{63.6±2.7}  & \underline{71.6±10.6} & 64.9±2.3        & 68.9±8.3         & 66.2±6.7         \\
\midrule
Decoupled + Reranking (Ours) & 0.0±0.0           & 32.0±2.2           & 64.4±6.7           & \underline{75.6±6.2}        & \underline{\textbf{78.2±7.7}}         & \underline{76.4±4.1}         \\ \bottomrule
\end{tabular}
}
\end{table}


To further examine the effect of actor-side policy improvement, we compare coupled and decoupled variants with and without critic-based reranking on Puzzle-4x4. \Cref{tab:exp:matched-proposer} reports the results across different candidate budgets $N$. Note that the variant \textit{Coupled} refers to the original version of FM-IQL, which outputs only a single action at each step and does not apply reranking. Therefore, it is not applicable to settings with $N\in\{2,4,8,16,32\}$, and these entries are marked as N/A in \Cref{tab:exp:matched-proposer}.
As shown, adding critic-based reranking to the coupled IQL policy (\emph{Coupled + Reranking}) consistently matching or outperforming the original IQL variant across all values of $N$, confirming that reranking itself can provide substantial gains.

More importantly, with an appropriate candidate budget, \emph{Decoupled + Reranking} further outperforms both the original IQL policy and its reranked counterpart, achieving the best overall performance across all methods and candidate budgets at $N=16$. These results indicate that reranking alone remains constrained by the actor learned under the coupled training paradigm, whereas the proposed decoupled policy extraction paradigm can better exploit critic-based selection and achieve stronger overall performance.


It is also worth noting that, \emph{Coupled + Reranking} reaches its best performance at $N=8$ and then degrades as $N$ further increases. As discussed in \cref{sec:distribution_shift}, one potential limitation of the coupled training paradigm is that the actor distribution can be distorted by critic estimation errors, leading to an OOD amplification loop. The observed degradation at larger candidate budgets provides further empirical support for this argument.

\subsection{Further Investigations}

\begin{figure}{}
    \centering
    \includegraphics[width=\textwidth]{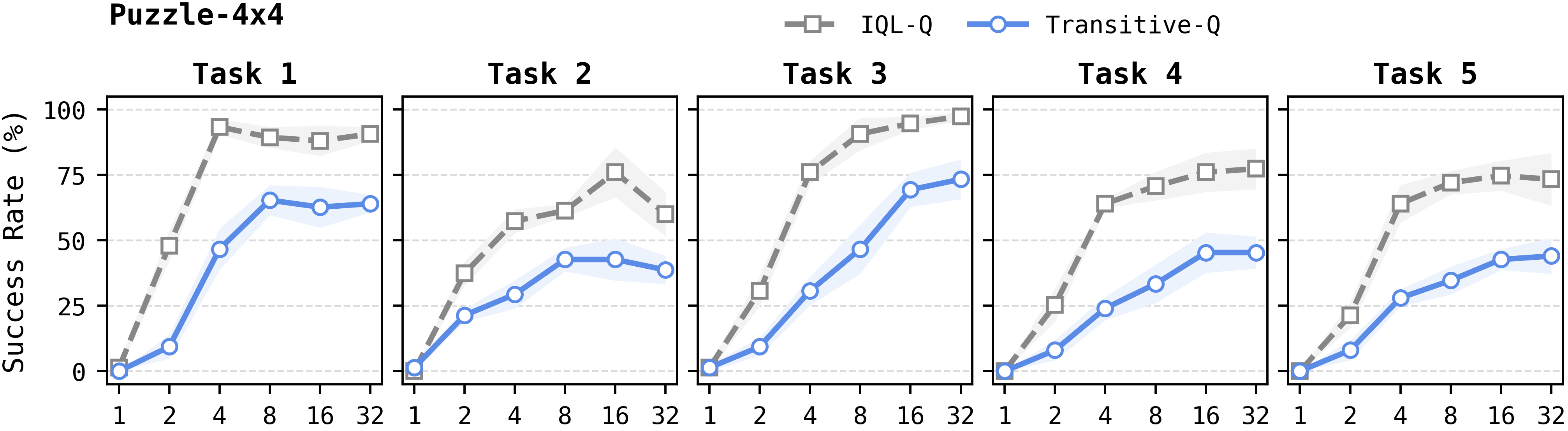}
    \caption{Impact of the number of candidates.}
    \label{fig:exp:impact of n}
\end{figure}

\textbf{Impact of the number of candidates $N$.}
Taking Puzzle as an example, we validate the impact of $N$ and summarize the impact in~\cref{fig:exp:impact of n}. As shown, performance initially improves as $N$ increases, but declines after reaching its peak.
We believe the reason is that, when $N$ is small, increasing $N$ allows the policy to explore the behavior-supported action region more thoroughly, thereby increasing the likelihood of identifying higher-value actions. However, as $N$ grows further, the expected number of generated out-of-distribution actions also increases. Since such actions are often erroneously assigned high values by the Q-function, they may be preferentially selected during reranking, ultimately leading to suboptimal decisions.

 \begin{table*}[!th]
  \centering
  \small
  \setlength{\tabcolsep}{5pt}
  \caption{Inference cost under different candidate budgets on the Cube-Double environment.}
  \label{tab:exp:inference_n}
   \begin{tabular}{lcrrrrr}
      \toprule
      Method
      & $N$
      & Success (\%)
      & Latency (ms)
      & Latency ratio
      & GFLOPs
      & GPU mem. (MiB) \\
      \midrule

      \multirow{2}{*}{FM-BC}
      & 1           & $7.20 \pm 2.00$          & 0.31 & 1.00$\times$ & 0.011 & 348 \\
      & 32 (random) & $6.67 \pm 1.46$          & 0.40 & 1.29$\times$ & 0.362 & 542 \\

      \midrule
      \multirow{6}{*}{IQL-Q}
      & 1  & $5.33 \pm 1.74$           & 0.33 & 1.09$\times$ & 0.011 & 356 \\
      & 2  & $69.33 \pm 1.12$          & 0.50 & 1.63$\times$ & 0.026 & 550 \\
      & 4  & $\mathbf{77.07 \pm 1.86}$ & 0.51 & 1.68$\times$ & 0.052 & 550 \\
      & 8  & $69.33 \pm 1.33$          & 0.50 & 1.64$\times$ & 0.105 & 550 \\
      & 16 & $68.80 \pm 2.97$          & 0.50 & 1.62$\times$ & 0.209 & 550 \\
      & 32 & $62.40 \pm 2.61$          & 0.44 & 1.44$\times$ & 0.434 & 552 \\

      \midrule
      \multirow{6}{*}{Transitive-Q}
      & 1  & $5.33 \pm 1.74$           & 0.33 & 1.07$\times$ & 0.011 & 356 \\
      & 2  & $19.47 \pm 1.61$          & 0.50 & 1.63$\times$ & 0.026 & 550 \\
      & 4  & $\mathbf{26.13 \pm 3.90}$ & 0.52 & 1.70$\times$ & 0.052 & 550 \\
      & 8  & $18.67 \pm 2.73$          & 0.50 & 1.64$\times$ & 0.105 & 550 \\
      & 16 & $11.73 \pm 1.07$          & 0.50 & 1.63$\times$ & 0.209 & 550 \\
      & 32 & $11.20 \pm 0.53$          & 0.44 & 1.43$\times$ & 0.434 & 552 \\

      \bottomrule
    \end{tabular}
  \end{table*}
\textbf{Inference cost.}
Our framework shifts critic-driven policy improvement from training-time actor
  optimization to inference-time candidate selection, which may introduce
  additional deployment cost as the candidate budget \(N\) increases.
  We study this trade-off on Cube-Double, a representative embodied manipulation
  task that requires time-sensitive online decision-making.
  At each decision step, all \(N\) candidates are generated in parallel through
  ten batched flow vector-field evaluations, followed by one batched twin-critic
  evaluation for reranking.
  Thus, increasing \(N\) widens the computation batch rather than adding serial
  proposer calls.
  As shown in \cref{tab:exp:inference_n}, although GFLOPs increase with \(N\),
  median latency remains below \(0.53\) ms across all configurations, while GPU
  memory remains nearly constant at \(550\)--\(552\) MiB for reranking budgets
  from \(N=2\) to \(N=32\).


%% file: sections/6.relatedworks.tex
\section{Related Works}
\label{sec:related_work}

\subsection{Actor-side policy improvement in offline RL}
Offline RL learns from a fixed dataset while deploying policies that may choose
actions different from those in the data, making extrapolation error a central
challenge \citep{levine2020offline}. A common response is
\emph{coupled policy improvement}: after learning a critic or value function,
the method trains a parameterized actor to stay close to the behavior
distribution while shifting probability toward actions assigned high value by
the critic. This paradigm underlies many offline RL methods, including
behavior-constrained or regularized approaches such as BCQ, BEAR, BRAC, and
TD3+BC \citep{fujimoto2019bcq,kumar2019bear,wu2019brac,fujimoto2021td3bc},
conservative or implicit value-learning methods such as CQL and IQL
\citep{kumar2020cql,kostrikov2022iql}, and actor-extraction objectives such as
AWR, AWAC, CRR, and One-step RL
\citep{peng2019awr,nair2021awac,wang2020crr,brandfonbrener2021onestep}. Recent
diffusion and flow-based offline RL methods further extend this actor-side view
by injecting critic information into expressive generative policies through
Q-maximization, energy guidance, distillation, or critic-gradient objectives
\citep{wang2023diffusionql,lu2023qgpo,park2025fql,alles2025flowq,li2026qam,chen2025modular}.

\subsection{Inference-time policy improvement}
Some other works perform policy improvement at inference time. For instance,
SfBC and EMaQ generate candidate actions from behavior models and select among
them using a critic~\citep{chen2023sfbc,ghasemipour2021emaq}. Yet they rely
on the BC actor for both proposal
generation and critic learning. In contrast, our paradigm only uses the
behavior model to generate candidate actions, reducing its influence on value
estimation.

Generative RL methods explore another form of inference-time improvement.
Diffuser guides trajectory generation using value models
\citep{janner2022planning}. QGF extends this idea to flow-matching policies
\citep{zhou2026test}; further, GAF applies QGF-style guidance to frozen
flow-matching VLA policies for embodied tasks~\citep{yang2026guided}. However,
action generation in QGF and GAF remains coupled with the critic during
inference. In contrast, our proposer generates candidates without
critic intervention, and the critic is applied only for post-generation
ranking.

A few prior works adopt similar decoupled structures, including V-GPS and IDQL
\citep{nakamoto2024vgps,hansenestruch2023idql}. Although similar, these methods focus on
specific applications or algorithmic instantiations rather than systematically
studying decoupling as a general policy extraction principle. We formalize this
design as a decoupled policy extraction framework and analyze its mechanisms
through controlled comparisons with coupled methods.



\blue{
}




\subsection{Diagnosing offline-RL bottlenecks}

Prior work studies which component of offline RL most limits performance.
Most relevantly, \citet{park2024bottleneck} identify policy extraction as a
major bottleneck. We build on this perspective by opening the policy-extraction
step itself: rather than treating extraction methods as monolithic rules, we
explicitly control the proposer, critic, selector, and candidate budget to
analyze when performance can be recovered through behavior-supported proposal
and critic-based selection.

%% file: sections/7.conclusion.tex
\section{Conclusion and Discussion}
In this work, we revisit whether learning the offline data distribution and policy improvement should be jointly carried out by the actor, and identify two fundamental limitations of coupled policy improvement offline RL methods, $i.e.$, critic-error amplification and the support--value trade-off.
To avoid these problems, we decouple policy improvement from actor training and decompose policy extraction into behavior-supported action proposal, critic evaluation, and inference-time selection. The results suggest that policy improvement is more effectively achieved by decoupling critic-based selection from actor training.

\paragraph{Implications for Future High-Capacity Policies.}
After decoupling policy improvement from actor learning, these expressive but parameter-intensive models need only be trained to model the dataset distribution through behavior cloning or supervised fine-tuning. Task-specific optimization can then be delegated to a lightweight critic that selects among actor-generated candidates at inference time, rather than repeatedly updating the actor using critic-derived objectives. This separation makes large pretrained policies substantially easier to incorporate into offline RL and allows a single proposal model to be reused across tasks by pairing it with different {cheaper} task-specific critics.



\paragraph{Limitations and Future Directions.}
The main limitation is proposal coverage. The critic can only select among
actions generated by the behavior-cloned proposer, whose distribution is
constrained by the offline dataset. Missing or extremely rare high-value
behaviors therefore impose a direct performance ceiling. Increasing the
candidate budget may improve coverage, but can also expose the selector to more
critic ranking errors and increases inference cost.

\section{Author Contributions}
\textbf{Xuyao Lin} led the experiments, prepared the visualizations, and contributed to analyzing the experimental results, writing the initial draft, and revising the manuscript. \\
\textbf{Yixiang Shan} conceived the study, designed the experiments, led the analysis and interpretation of the experimental results, formulated the main conclusions, and led the writing of the initial draft, as well as the subsequent review and revision of the manuscript. \\
\textbf{Jinru Duan} prepared the visualizations and contributed to writing the initial draft and reviewing and revising the manuscript. \\
\textbf{Tao Yang} developed the methodology, contributed to implementing the method and conducting the experiments, and participated in writing the initial draft and reviewing and revising the manuscript. \\
\textbf{Xinyu Zhao, Runyu Lei, Yiming Zhao, and Jiaxin Fan} contributed to reviewing and revising the manuscript. \\
\textbf{Zongbao Feng and Peng Jia} supervised the research and contributed to reviewing and revising the manuscript.

%% file: sections/8.appendix.tex
\section{Appendix}
\subsection{Implementation Details}
\label{sec:appendix:Implementation Details}
The optimization, architecture, goal-sampling, and evaluation settings used in
our experiments are summarized below.  Unless noted otherwise, each setting is
shared across methods and environments.

We evaluate six standard OGBench environments with oracle goal representations.  We compare a flow-matching behavior-cloning policy (FM-BC), three frozen-policy rerankers (BC$+$IQL-Q, BC$+$Q-Learning-Q, and BC$+$Transitive-Q), FM-IQL, and the official TRL baseline using reparameterized policy gradients (TRL-RPG).  

 For the standard OGBench benchmark, each training seed is evaluated at
  800K, 900K, and 1M gradient updates. The frozen FM-BC and BC+Q variants use five paired training seeds: critic
  seed \(i\) is evaluated with the FM-BC proposal trained using seed \(i\).

\subsection{Hyperparameter Settings}
\label{sec:appendix:hyperparameters}

Each trainable component is optimized for \(10^6\) gradient steps.  For each BC$+$Q method, we first train an FM-BC proposal policy for \(10^6\) steps and subsequently freeze it while training the corresponding critic for an additional \(10^6\) steps. In contrast, FM-IQL and TRL-RPG jointly optimize their actor and critic for \(10^6\) steps.

Following the notation used by OGBench and TRL, goal-sampling ratios are
  reported as
  \[
      \rho =
      (p_{\mathrm{cur}},p_{\mathrm{geom}},
       p_{\mathrm{traj}},p_{\mathrm{rand}}),
  \]
  where \(p_{\mathrm{geom}}\) and \(p_{\mathrm{traj}}\) denote geometrically
  and uniformly sampled future trajectory goals, respectively.

\begin{table}[H]
      \centering
      \footnotesize
      \setlength{\tabcolsep}{4pt}
      \renewcommand{\arraystretch}{1.08}
      \caption{Hyperparameters shared across the standard OGBench
      experiments.}
      \label{tab:ogbench-common-hparams}
      \begin{tabular}{@{}p{0.36\linewidth}p{0.58\linewidth}@{}}
          \toprule
          Hyperparameter & Value \\
          \midrule
          Gradient steps per trained model
              & \(10^6\) \\
          Optimizer
              & Adam \\
          Learning rate
              & \(3\times10^{-4}\) \\
          Batch size
              & \(1024\) \\
          Actor hidden dimensions
              & \([512,512,512]\) \\
          Critic/value hidden dimensions
              & \([512,512,512]\) \\
          Nonlinearity
              & GELU \\
          Layer normalization
              & enabled for actor and critic/value networks \\
          Critic ensemble size
              & \(2\) \\
          Target-network update rate \(\tau\)
              & \(0.005\) \\
          Action range
              & \([-1,1]\) \\
          Actor goal ratio \(\rho_{\pi}\)
              & \((0,0,1,0)\) \\
          IQL/Q-learning goal ratio \(\rho_Q\)
              & \((0.2,0.5,0,0.3)\) \\
          Transitive goal ratio \(\rho_{\mathrm{tr}}\)
              & \((0,1,0,0)\) \\
          Sparse reward for Bellman critics
              & \(0\) when the sampled goal is the current state,
                and \(-1\) otherwise \\
          \bottomrule
      \end{tabular}
  \end{table}

 \begin{table}[H]
      \centering
      \scriptsize
      \setlength{\tabcolsep}{3pt}
      \renewcommand{\arraystretch}{1.10}
      \caption{Environment- and task-specific hyperparameters.
      The vectors contain \((N_1,\ldots,N_5)\) in the task order used by
      the corresponding OGBench environment.  For BC$+$IQL-Q and
      BC$+$Transitive-Q, \(N\) is selected independently for every method
      and task from \(\{1,2,4,8,16,32\}\) using the five-seed mean.}
      \label{tab:ogbench-task-hparams}
      \begin{tabular*}{\linewidth}
          {@{\extracolsep{\fill}}lcccccc@{}}
          \toprule
          Environment
              & \(\gamma\)
              & \(\alpha_{\mathrm{FM}}\)
              & \(\lambda_{\mathrm{tr}}\)
              & \shortstack{BC$+$IQL-Q\\\(N_{1:5}\)}
              & \shortstack{BC$+$Transitive-Q\\\(N_{1:5}\)}\\
          \midrule
          AntMaze-Large
              & \(0.99\)
              & \(1.0\)
              & \(0.0\)
              & \((16,1,2,1,1)\)
              & \((32,1,2,2,2)\)\\
          HumanoidMaze-Med.
              & \(0.995\)
              & \(0.4\)
              & \(0.0\)
              & \((16,8,2,1,2)\)
              & \((16,8,4,4,8)\) \\
          AntSoccer-Arena
              & \(0.99\)
              & \(0.4\)
              & \(0.5\)
              & \((8,8,4,8,8)\)
              & \((16,8,8,16,4)\) \\
          Cube-Double
              & \(0.99\)
              & \(2.5\)
              & \(1.0\)
              & \((4,4,4,16,4)\)
              & \((4,4,4,4,2)\)\\
          Scene-Play
              & \(0.99\)
              & \(2.5\)
              & \(1.0\)
              & \((2,8,4,8,2)\)
              & \((8,8,8,16,16)\) \\
          Puzzle-4x4
              & \(0.99\)
              & \(2.5\)
              & \(2.0\)
              & \((4,16,32,32,16)\)
              & \((8,8,32,16,32)\) \\
          \bottomrule
      \end{tabular*}
  \end{table}

%
\subsection{CRL and QRL Baselines}
\label{sec:appendix:crl-qrl}

We additionally evaluate the original coupled actors of contrastive
reinforcement learning (CRL) and quasimetric reinforcement learning (QRL) on
the same six OGBench oracle-representation environments.  We use the
environment-specific settings from the official OGBench implementation
\citep{park2025ogbench}, together with the oracle-representation adapter
described below. The complete results are
shown in \cref{tab:appendix:crl-qrl-results}. The BC coefficient \(\alpha_{\mathrm{BC}}\) below scales the negative
log-likelihood term in the coupled actor objective.  QRL uses an interval
quasimetric embedding (IQE).

\begin{table*}[h]
    \centering
    \scriptsize
    \setlength{\tabcolsep}{5pt}
    \renewcommand{\arraystretch}{0.90}
    \caption{CRL and QRL success rates.  Each seed-level value first averages the 800K, 900K,
    and 1M checkpoints.}
    \label{tab:appendix:crl-qrl-results}
    \begin{tabular*}{\textwidth}{@{\extracolsep{\fill}}llcc@{}}
        \toprule
        Environment & Subtask & CRL & QRL \\
        \midrule
        \multirow{6}{*}{AntMaze-Large}
            & Task 1 & \(87.56 \pm 3.28\) & \(56.44 \pm 6.49\) \\
            & Task 2 & \(87.11 \pm 3.50\) & \(69.11 \pm 4.63\) \\
            & Task 3 & \(87.78 \pm 3.71\) & \(92.44 \pm 2.27\) \\
            & Task 4 & \(88.00 \pm 2.49\) & \(55.33 \pm 8.02\) \\
            & Task 5 & \(89.78 \pm 1.37\) & \(61.56 \pm 5.48\) \\
            & \textbf{overall} & \(88.04 \pm 1.49\) & \(66.98 \pm 3.46\) \\
        \midrule
        \multirow{6}{*}{HumanoidMaze-Medium}
            & Task 1 & \(91.56 \pm 2.45\) & \(13.56 \pm 5.17\) \\
            & Task 2 & \(91.78 \pm 1.66\) & \(38.89 \pm 0.31\) \\
            & Task 3 & \(58.89 \pm 24.91\) & \(31.78 \pm 8.95\) \\
            & Task 4 & \(5.11 \pm 4.16\) & \(11.78 \pm 3.46\) \\
            & Task 5 & \(95.33 \pm 1.09\) & \(31.33 \pm 2.88\) \\
            & \textbf{overall} & \(68.53 \pm 4.49\) & \(25.47 \pm 3.71\) \\
        \midrule
        \multirow{6}{*}{AntSoccer-Arena}
            & Task 1 & \(49.78 \pm 4.01\) & \(16.89 \pm 3.33\) \\
            & Task 2 & \(34.00 \pm 4.11\) & \(17.33 \pm 1.09\) \\
            & Task 3 & \(56.44 \pm 5.37\) & \(21.33 \pm 1.44\) \\
            & Task 4 & \(14.89 \pm 2.06\) & \(7.11 \pm 2.51\) \\
            & Task 5 & \(24.22 \pm 1.66\) & \(8.89 \pm 2.45\) \\
            & \textbf{overall} & \(35.87 \pm 2.94\) & \(14.31 \pm 0.79\) \\
        \midrule
        \multirow{6}{*}{Cube-Double}
            & Task 1 & \(70.67 \pm 5.76\) & \(6.22 \pm 0.83\) \\
            & Task 2 & \(47.33 \pm 1.44\) & \(0.22 \pm 0.31\) \\
            & Task 3 & \(41.78 \pm 2.45\) & \(0.00 \pm 0.00\) \\
            & Task 4 & \(0.67 \pm 0.54\) & \(0.22 \pm 0.31\) \\
            & Task 5 & \(22.22 \pm 1.26\) & \(0.00 \pm 0.00\) \\
            & \textbf{overall} & \(36.53 \pm 1.15\) & \(1.33 \pm 0.11\) \\
        \midrule
        \multirow{6}{*}{Scene-Play}
            & Task 1 & \(64.89 \pm 3.10\) & \(14.67 \pm 3.27\) \\
            & Task 2 & \(17.78 \pm 4.53\) & \(0.67 \pm 0.54\) \\
            & Task 3 & \(40.22 \pm 3.71\) & \(2.22 \pm 1.13\) \\
            & Task 4 & \(5.56 \pm 1.66\) & \(2.00 \pm 0.00\) \\
            & Task 5 & \(0.89 \pm 1.26\) & \(0.44 \pm 0.63\) \\
            & \textbf{overall} & \(25.87 \pm 0.61\) & \(4.00 \pm 0.58\) \\
        \midrule
        \multirow{6}{*}{Puzzle-4x4}
            & Task 1 & \(0.22 \pm 0.31\) & \(0.44 \pm 0.63\) \\
            & Task 2 & \(0.67 \pm 0.00\) & \(0.00 \pm 0.00\) \\
            & Task 3 & \(0.00 \pm 0.00\) & \(0.22 \pm 0.31\) \\
            & Task 4 & \(0.00 \pm 0.00\) & \(0.00 \pm 0.00\) \\
            & Task 5 & \(0.22 \pm 0.31\) & \(0.22 \pm 0.31\) \\
            & \textbf{overall} & \(0.22 \pm 0.06\) & \(0.18 \pm 0.25\) \\
        \midrule
        \multicolumn{2}{@{}l}{\textbf{Six-environment macro average}}
            & \(42.51 \pm 1.14\) & \(18.71 \pm 1.00\) \\
        \bottomrule
    \end{tabular*}
\end{table*}

\begin{table}[t]
    \centering
    \footnotesize
    \setlength{\tabcolsep}{5pt}
    \renewcommand{\arraystretch}{1.05}
    \caption{Environment-specific CRL and QRL hyperparameters.}
    \label{tab:appendix:crl-qrl-env-hparams}
    \begin{tabular}{@{}lccc@{}}
        \toprule
        Environment & \(\gamma\) & CRL \(\alpha_{\mathrm{BC}}\)
            & QRL \(\alpha_{\mathrm{BC}}\) \\
        \midrule
        AntMaze-Large & \(0.99\) & \(0.1\) & \(0.003\) \\
        HumanoidMaze-Medium & \(0.995\) & \(0.1\) & \(0.001\) \\
        AntSoccer-Arena & \(0.99\) & \(0.3\) & \(0.003\) \\
        Cube-Double & \(0.99\) & \(3.0\) & \(0.3\) \\
        Scene-Play & \(0.99\) & \(3.0\) & \(0.3\) \\
        Puzzle-4x4 & \(0.99\) & \(3.0\) & \(0.3\) \\
        \bottomrule
    \end{tabular}
\end{table}

\subsection{Actor-Drift Diagnostic: Implementation Details}
  \label{sec:appendix:actor-drift-hparams}

  This appendix provides the complete actor objectives and reproducibility
  details for the actor-drift diagnostic in
  \cref{fig:exp:distribution_shift}. All actions
  shown in the diagnostic are generated directly by the corresponding diffusion
  actor, without candidate reranking or \(Q\)-dependent inference.

  \paragraph{Actor objectives.}
  All variants use the same diffusion actor and behavior-cloning loss
  \(\mathcal{L}_{\mathrm{diff}}\). Their objectives are
  \begin{equation}
      \mathcal{L}_{\mathrm{actor}}
      =
      \mathcal{L}_{\mathrm{diff}}
      +
      \lambda\mathcal{L}_{\mathrm{guide}},
  \end{equation}
  where
  \begin{align}
      \mathcal{L}_{\mathrm{guide}}^{\mathrm{CRR}}
      &=
      -\mathbb{E}\!\left[
          \operatorname{sg}\!\left(w(A_{\mathrm{CRR}})\right)
          \log \mathcal{N}(a;\hat a_\theta,I)
      \right], \\
      \mathcal{L}_{\mathrm{guide}}^{\mathrm{TD3}}
      &=
      -\mathbb{E}\!\left[
          \frac{
              \alpha_{\mathrm{TD3}}
              Q_j(s,\hat a_\theta,g)
          }{
              \operatorname{sg}\!\left(
                  \mathbb{E}_{\mathcal B}
                  \left|Q_j(s,\hat a_\theta,g)\right|
              \right)
          }
      \right].
  \end{align}
  Here, \(\hat a_\theta\) is the reconstructed clean action,
  \(j\sim\operatorname{Uniform}\{1,2\}\), and \(\operatorname{sg}\) denotes
  stop-gradient. For CRR,
  \[
      w(A)=\min\{w_{\max},\exp(A/\beta)\},
  \]
  where \(A_{\mathrm{CRR}}\) compares the dataset-action value against a
  policy-sampled value baseline. EDP-BC sets
  \(\mathcal{L}_{\mathrm{guide}}=0\).

  \paragraph{PCA and BC95 construction.}
  A separate BC reference cloud is generated from an independent set of \(1000\)
  initial noise vectors, which is reused across checkpoints. At each checkpoint,
  the first \(500\) projected BC actions form the KDE-fit split and the remaining
  \(500\) form the conformal-calibration split. The KDE-fit actions from all four
  checkpoints are pooled to fit one shared PCA basis.

  For each checkpoint, we fit a two-dimensional Gaussian KDE with Scott
  bandwidth and full covariance. The nonconformity score is the negative log KDE
  density. With \(500\) calibration actions, the BC95 threshold is the
  \(476\)-th smallest calibration score,
  \(\lceil(500+1)\times0.95\rceil=476\). An action is classified as outside BC95
  when its split-conformal \(p\)-value is at most \(0.05\). This classification
  measures coverage relative to the same-checkpoint BC distribution in the
  shared PCA plane and is not a formal environment-level OOD test.

  \paragraph{Visual encoding.}
  All panels use the same PCA basis, but their axis limits are adjusted
  independently to contain the complete action cloud and BC95 contour. Points
  inside and outside BC95 use identical size, opacity, color mapping, and draw
  order. The dashed BC95 contour is rendered above the action samples.

  CRR actions are colored by the diagnostic advantage
  \[
  A_{\mathrm{diag}}(s,a,g)
  =
  Q_{\min}(s,a,g)
  -
  \frac{1}{K}\sum_{k=1}^{K}
  Q_{\min}(s,\tilde a_k,g),
  \]
  where \(\tilde a_k\) are sampled from the current CRR policy. One robust color
  range is shared across all CRR checkpoints. TD3 actions are colored by the
  within-checkpoint percentile rank of \(Q_{\min}=\min(Q_1,Q_2)\); consequently,
  TD3 colors indicate within-panel rankings and should not be compared as
  absolute \(Q\)-values across checkpoints.

  \begin{table}[H]
      \centering
      \footnotesize
      \setlength{\tabcolsep}{4pt}
      \renewcommand{\arraystretch}{1.08}
      \caption{Hyperparameters for the diffusion-policy actor-drift diagnostic
      on AntSoccer-Arena.}
      \label{tab:actor-drift-hparams}
      \begin{tabular}{@{}p{0.36\linewidth}p{0.58\linewidth}@{}}
          \toprule
          Hyperparameter & Value \\
          \midrule
          \multicolumn{2}{@{}l}{\textit{Shared training settings}} \\
          \addlinespace[2pt]
          Environment
              & antsoccer-arena-navigate-oraclerep-v0 \\
          Gradient steps
              & \(10^6\) per method \\
          Optimizer
              & AdamW \\
          Learning rate
              & \(3\times10^{-4}\) \\
          Batch size
              & \(1024\) \\
          Diffusion actor
              & \(3\times512\) MLP, Mish activation, LayerNorm enabled \\
          Actor-goal probabilities
              & \((p_{\mathrm{cur}},p_{\mathrm{traj}},p_{\mathrm{rand}})
                 =(0,1,0)\), uniform trajectory-goal sampling \\
          Diffusion timesteps / schedule
              & \(100\) / linear \\
          Time-embedding dimension
              & \(16\) \\
          Inference sampler
              & deterministic DDIM, \(15\) steps, \(\eta=0\) \\
          Diffusion-loss coefficient
              & \(c_{\mathrm{diff}}=1.0\) \\
          Action range
              & \([-1,1]\) \\
          \midrule
          \multicolumn{2}{@{}l}{\textit{Coupled-critic settings: CRR and TD3 only}} \\
          \addlinespace[2pt]
          Critic
              & double \(Q\), \(3\times512\) MLP \\
          Discount factor
              & \(\gamma=0.99\) \\
          Target-network update
              & \(\tau=0.005\), every \(5\) updates after update \(1000\) \\
          Critic-goal probabilities
              & \((p_{\mathrm{cur}},p_{\mathrm{traj}},p_{\mathrm{rand}})
                 =(0.2,0.5,0.3)\), uniform trajectory-goal sampling \\
          Sparse critic reward
              & \(0\) for the current-state goal and \(-1\) otherwise \\
          \midrule
          \multicolumn{2}{@{}l}{\textit{Method-specific settings}} \\
          \addlinespace[2pt]
          EDP-BC
              & no critic or guide loss \\
          EDP-CRR
              & \(\beta=15\), \(w_{\max}=20\), \(K=10\),
                baseline \(\sigma=0.5\), \(c_{\mathrm{guide}}=0.3\),
                critic LayerNorm enabled \\
          EDP-TD3
              & \(\alpha_{\mathrm{TD3}}=1.5\),
                \(c_{\mathrm{guide}}=1.0\),
                critic LayerNorm disabled \\
          \bottomrule
      \end{tabular}
  \end{table}

\subsection{CQL Conservatism Diagnostic}
  \label{sec:appendix:cql-tradeoff}

  This appendix describes the goal-conditioned CQL diagnostic reported in
  \cref{fig:exp:claim2}. Experiments are conducted on
  \texttt{cube-double-play-oraclerep-v0} using a tanh-Gaussian actor and a
  double critic. The critic augments the standard Bellman objective with an
  importance-corrected CQL log-sum-exp penalty weighted by
  \(\alpha_{\mathrm{CQL}}\). The penalty uses \(10\) samples from the uniform
  action distribution, the current policy, and the policy at the next
  observation. The SAC actor objective is held fixed across
  \(\alpha_{\mathrm{CQL}}\), so conservatism affects the policy only through the
  learned critic.

  \paragraph{BC95 construction.}
  For each repeat, we uniformly sample \(32\) valid dataset trajectories
  without replacement. Each anchor consists of the midpoint nonterminal state
  of one trajectory and its terminal oracle goal. We generate \(1000\) actions
  from the evaluated CQL actor and an independent \(1000\)-action reference
  cloud from a behavior-cloning actor trained for \(10^6\) updates.

  For every anchor, the first \(500\) BC actions fit a two-dimensional PCA
  projection and a Gaussian KDE with Scott bandwidth. The remaining \(500\)
  actions form the split-conformal calibration set. Using negative log KDE
  density as the nonconformity score \(R\).
  An action is counted as outside BC95 when \(p(a)\leq0.05\).
  This procedure is repeated five times using matched anchor and sampling seeds.
  Outside BC95 measures deviation from a same-state BC reference in the
  two-dimensional PCA plane; it is not a formal environment-level OOD test.

  \paragraph{Training and aggregation.}
  All policies are trained for \(10^6\) gradient updates. Success is evaluated
  deterministically with evaluation seed \(0\). When available, the
  \(800\mathrm{K}\), \(900\mathrm{K}\), and \(1\mathrm{M}\) checkpoints are
  first averaged within each model. Outside BC95 is averaged over anchors,
  the five conformal repeats, selected checkpoints, and finally the available
  model slots.

  \begin{table}[t]
      \centering
      \footnotesize
      \setlength{\tabcolsep}{4pt}
      \renewcommand{\arraystretch}{1.08}
      \caption{Shared settings for the CQL conservatism diagnostic.}
      \label{tab:appendix:cql-tradeoff}
      \begin{tabular}{@{}p{0.35\linewidth}p{0.59\linewidth}@{}}
          \toprule
          Setting & Value \\
          \midrule
          Environment
              & \texttt{cube-double-play-oraclerep-v0} \\
          Gradient updates
              & \(10^6\) \\
          Optimizer / learning rate
              & Adam / \(3\times10^{-4}\) \\
          Batch size
              & \(1024\) \\
          Actor
              & \(3\times512\) MLP, LayerNorm, tanh output \\
          Critic
              & Double \(Q\), \(3\times512\) MLP, LayerNorm \\
          Discount / target update
              & \(\gamma=0.99\), \(\tau=0.005\) \\
          CQL temperature
              & \(1.0\) \\
          Proposal samples
              & \(10\) per proposal source \\
          Actor goal probabilities
              & \((0,1,0)\) \\
          Critic goal probabilities
              & \((0.2,0.5,0.3)\) \\
          \bottomrule
      \end{tabular}
  \end{table}